\documentclass[letterpaper, 10 pt, conference]{ieeeconf}  

\IEEEoverridecommandlockouts                              

\usepackage{graphicx} 
\usepackage{subfig}
\usepackage{url}
\usepackage{array}
\usepackage{geometry}
\newcolumntype{I}{!{\vrule width 1pt}}
\newlength\savedwidth
\newcommand\whline{\noalign{\global\savedwidth\arrayrulewidth
		\global\arrayrulewidth 1pt}%
	\hline
	\noalign{\global\arrayrulewidth\savedwidth}}
\newlength\savewidth
\newcommand\shline{\noalign{\global\savewidth\arrayrulewidth
		\global\arrayrulewidth 0.05pt}
	\hline
	\noalign{\global\arrayrulewidth\savewidth}}
\usepackage{color}

\title{\LARGE \bf
Dilated Continuous Random Field for Semantic Segmentation
}

\author{Xi Mo$^{1, *}$, Xiangyu Chen$^{1}$, Cuncong Zhong$^{1}$, Rui Li$^{2}$, Kaidong Li$^{1}$, Usman Sajid$^{1}$
\thanks{*Corresponding Author, email: x618m566@ku.edu}
\thanks{$^{1}$School of Engineering, University of Kansas, Lawrence KS 66049, USA}%
\thanks{$^{2}$Wetland Studies and Solutions, Inc., Virginia Beach VA 23451, USA}%
}

\begin{document}

\maketitle
\thispagestyle{empty}
\pagestyle{empty}

\begin{abstract}

Mean field approximation methodology has laid the foundation of modern Continuous Random Field (CRF) based solutions for the refinement of semantic segmentation. In this paper, we propose to relax the hard constraint of mean field approximation -  minimizing the energy term of each node from probabilistic graphical model, by a  global optimization with the proposed dilated sparse convolution module (DSConv). In addition, adaptive global average-pooling and adaptive global max-pooling are implemented as replacements of fully connected layers. In order to integrate DSConv, we design an end-to-end, time-efficient DilatedCRF pipeline. The unary energy term is derived either from pre-softmax and post-softmax features, or the predicted affordance map using a conventional classifier, making it easier to implement DilatedCRF for varieties of classifiers. We also present superior experimental results of proposed approach on the suction dataset comparing to other CRF-based approaches.

\end{abstract}

\section{\MakeUppercase{Introduction}}\label{sec:intro}

\begin{figure}[t]
	\centering
	\includegraphics[width=\linewidth]{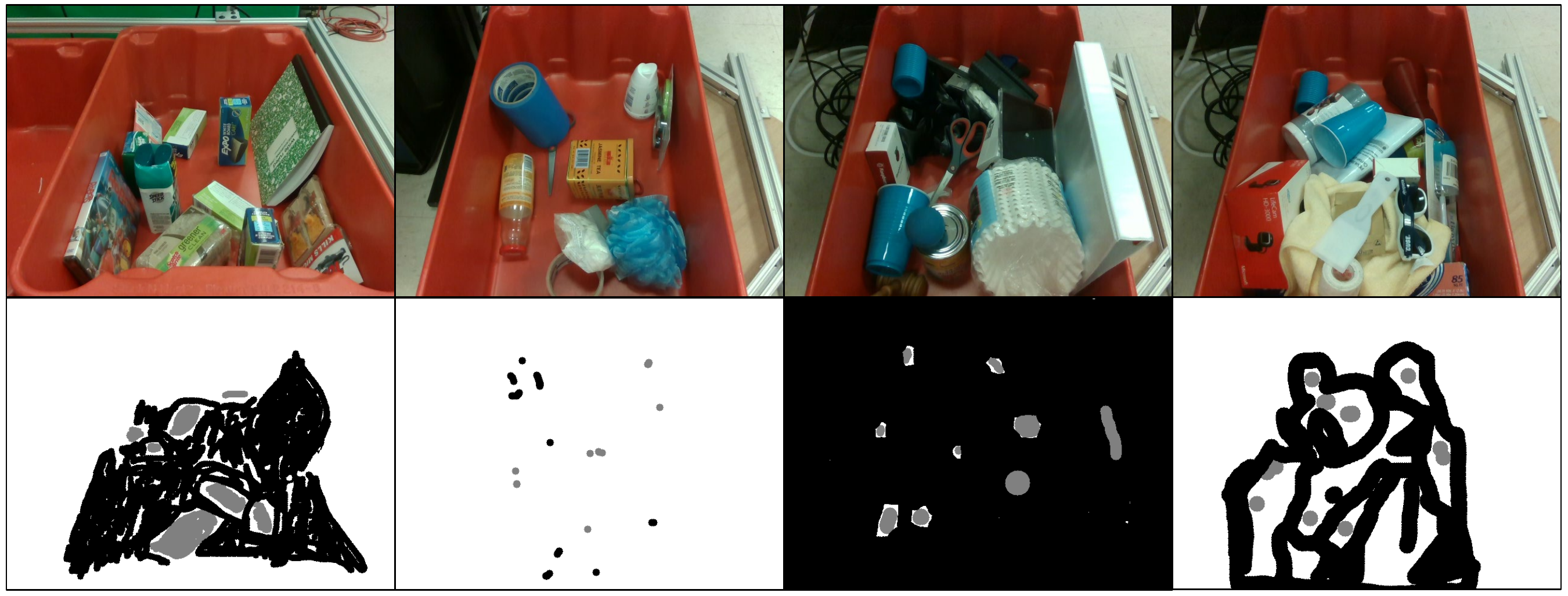}
	\caption{Different styles of affordance maps. Raw images (upper) and ground-truths (lower) are selected from the suction dataset~\cite{suctiondataset}, black regions signify the negative samples, grey regions the valid spots for suction, white regions the background.}
	\label{fig:1}
\end{figure}

An affordance map indicates valid areas for suction-based bin-picking, ignored background, and negative samples that signify an object w/o valid suction areas, as shown in Fig.~\ref{fig:1}. The annotations are labeled by multiple human experts, which is not precise enough due to several reasons: the same object is annotated in various styles by different experts; annotations are yielded via seemingly paradoxical inferences, e.g., a glass is pickable while a plastic ball is not; some objects, such as a laser pen, or largely occluded objects, are labeled as pickable in some affordance maps, whereas being unpickable in others. 

Therefore, the task to predict precise suction area is challenging given such conditions. No matter which approach is implemented to solve this challenging semantic segmentation task, the goal of CRF based post-processing is to refurbish the predictions into more fine-grained artifacts. As to the post-processing itself, real-time performance is preferred for robotic applications.

As one of the most popular works in demonstrating CRF for the field of semantic segmentation, the DenseCRF~\cite{krahenbuhl2011efficient}, proposes mean field approximation theory to approximate global energy minimization by optimizing local unary and pairwise potentials that encode relative positional and color information of entire image, and the message passing from other features to the local feature becomes the computational bottleneck, especially when the image is large in size. DenseCRF proposes to further boost the efficiency by lattice permutation~\cite{adams2010fast}, which reduces the computational complexity from $O(N^2)$ to $O(N)$. Still, it remains to be a key topic seeking the perfect trade-off between better global approximation and efficiency.

In this paper, we investigate the implementation of standard convolution upon CRF, and decreasing the usage of fully-connect layers. It seems natural that single linear layer can learn relative relationship between a local feature and all other features by connecting all these features and output one pairwise energy term. Suppose a feature volume with a height and width of 60 and 80, this linear layer has a total trainable parameters of 4.8K. Consider all features of the feature volume, the total number of parameters is 23.04M, which will be subjected to quadratic growth if its height and width double. As a remedy to reduce total parameters, we propose the dilated sparse convolution (DSConv) module that performs fast channel-wise convolution with a static kernel, then extract global energy terms using adaptive global max pooling (AGMP) and adaptive global average pooling (AGAP) as replacements of linear layers. With DSConv modules, Dilated CRF decreases 23.04M parameters to 0.36M, while showing state-of-the-art performance.

DSConv module correlates features in a global manner other than following the mean field approximation methodology. Since DilatedCRF is end-to-end trained with a GPU, it can learn trainable parameters globally without customized CPU-only operations, e.g., the computation of potentials with L2-norms and convolution optimized by lattice permutation~\cite{krahenbuhl2011efficient, zheng2015conditional}, patch-wise multiplication~\cite{teichmann2019convolutional},  etc. Also, it should be noted that, the massive concatenation of output features, and memory transfer consume the majority of runtime. One of our future works is to accelerate these trivial operations.

\begin{figure*}[t]
	\centering
	\includegraphics[width=\linewidth]{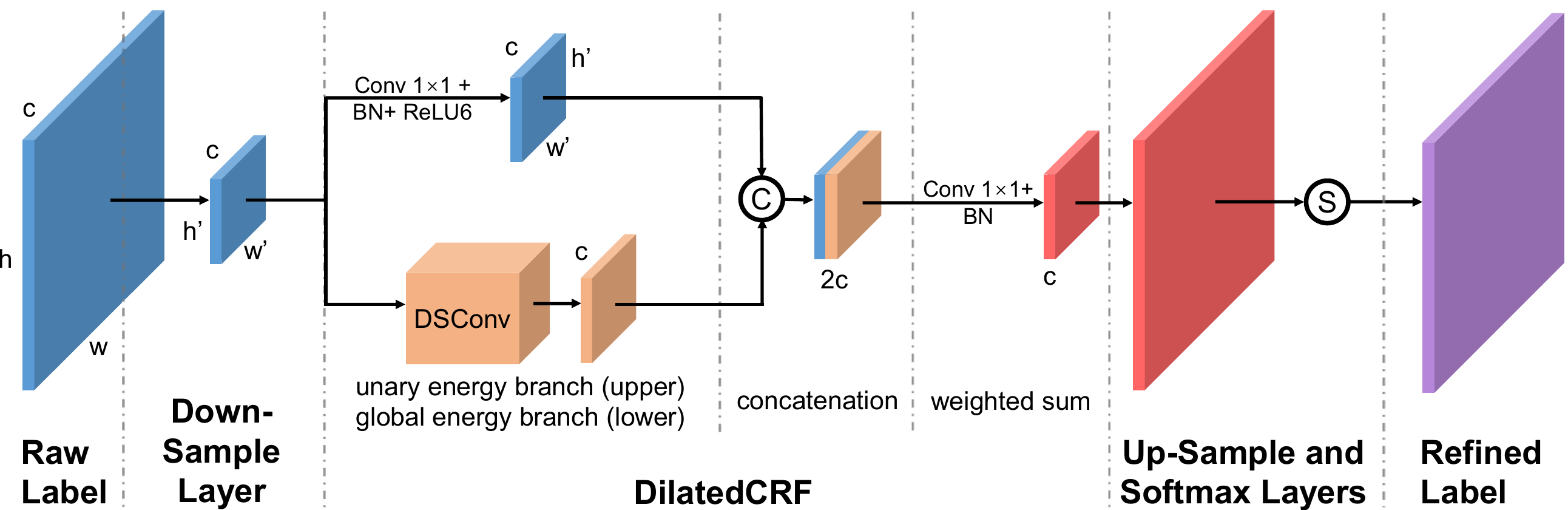}
	\caption{DilatedCRF pipeline. `Raw label' signifies the predicted affordance map or raw feature volume since all of them encode the unary energy term presented in (\ref{eqn:7}). The forward passing has three stages - down-sampling by a manually specified scale (multipliers 1/8, 1/4 and 1/2 for the suction dataset~\cite{suctiondataset}), DilatedCRF, and up-sampling to original label sizes, the final softmax operation interprets Gibbs energy (refer to (\ref{eqn:3})) optimization as minimizing the KL-Divergence between refined label and one-hot encoded ground-truth\protect\footnotemark[3].}
	\label{fig:2}
\end{figure*}

\section{\MakeUppercase{Continuous Random Field}}

\subsection{The General Form}

Suppose a label space $\mathcal{L}=\{l_i\},i=1,...,h\times w$, $h$ and $w$ are the height and width of the ground-truth, random variable $l_i\in\{y_i\},i=1,2,3$, $y_i$ corresponds to a category of the suction dataset~\cite{suctiondataset}, and input feature volume $\mathcal{X}=\{\mathbf{x}_j\}, j=1,...,h\times w$, each $\mathbf{x}_i$ is a 3-D feature vector. According to Hammersley-Clifford theorem, the CRF defined by $\mathcal{L}$ and $\mathcal{X}$ is then denoted as a Gibbs distribution over the set $\mathcal{C}$ of all maximal cliques:
\begin{equation}\label{eqn:1}
	P(\mathcal{L}|\mathcal{X}) = \frac{\prod_{c\in \mathcal{C}}\psi_{c}(\mathcal{L}|\mathcal{X})}{Z(\mathcal{X})},
\end{equation}
potential $\psi_{c}(\mathcal{L}|\mathcal{X})$ signifies a strict positive function defined on maximal clique $c$ of undirect graph $G$. In the context of semantic segmentation, let $V$ be the set of all spatial locations, $E$ the edges that connect vertices from $V$, then $G=(V, E)$. $Z(\mathcal{X})$ is the normalization factor that ensures $P(\mathcal{L}|\mathcal{X})$ a valid probabilistic distribution,
\begin{equation}\label{eqn:2}
	Z(\mathcal{X})=\sum_{\mathcal{L}}\prod_{c\in \mathcal{C}}\psi_{c}(\mathcal{L}|\mathcal{X}).
\end{equation} 

CRF refinement is achieved by the maximum a posteriori (MAP) labeling $\mathcal{L}^* = \arg \max_{\mathcal{L}} P(\mathcal{L}|\mathcal{X})$ w.r.t. ground-truth.

\footnotetext[3]{Specifically, the final softmax layer encapsules Gibbs energy term $E(\mathcal{L}|\mathcal{X})$ without the negative sign shown in (\ref{eqn:3}), indicating the tendency of neural network to learn the maximal probability distribution $P(\mathcal{L}^*|\mathcal{X})$ via gradients back-propagation, whereas how to reach the optimal $\mathcal{L}^*$ remains to be solved.}

Complete graph $G$ is ideal for refinement. However, its defect is also noteworthy - massive computation to exhaustively enumerate all possible combinations of random variable $\mathcal{L}$. To avoid the prod operation, exponential function is applied to potentials such that
\begin{equation}\label{eqn:3}
	P(\mathcal{L}|\mathcal{X}) = \frac{\exp(-E(\mathcal{L}|\mathcal{X}))}{Z(\mathcal{X})},
\end{equation}
where the Gibbs energy $E(\mathcal{L}|\mathcal{X}) = \sum_{c\in \mathcal{C}}\psi_{c}(\mathcal{L}|\mathcal{X})$. According to (\ref{eqn:3}), optimization of a CRF can be denoted as $\mathcal{L}^*=\arg\min_{\mathcal{L}}E(\mathcal{L}|\mathcal{X})$, $Z(\mathcal{\mathcal{X}})$ is ignored because it only depends on input feature volume $\mathcal{X}$.

\subsection{DenseCRF for Semantic Segmentation}

Dense CRF employs a fully-connected graph $G$ such that there's only one maximal clique. Let $N=h\times w$, the Gibbs energy has the form
\begin{equation}\label{eqn:4}
	E(\mathcal{L}|\mathcal{X}) = \sum_{1\leq i\leq N} \psi_{u}(l_i|\mathcal{X}) + \sum_{1\leq i<j\leq N} \psi_{p}(l_i, l_j|\mathcal{X}),
\end{equation}
where $\psi_{u}$, $\psi_{p}$ are unary and pairwise potentials respectively. Commonly-implemented DenseCRF~\cite{krahenbuhl2011efficient} defines  $\psi_{u}(\mathcal{L}|\mathcal{X})$ as the output of a classifier given input $\mathcal{X}$, and $\psi_{p}$ as
\begin{equation}\label{eqn:5}
	\psi_{p} (l_i, l_j|\mathcal{X}) = \mu(l_i, l_j)\sum_{m=1}^{M} w^{(m)} k^{(m)}(\mathbf{x}_i, \mathbf{x}_j),
\end{equation} 
where $\mu(l_i, l_j)$ is the compatibility function which is manually assigned by Potts model $\mu (l_i, l_j)=[l_i\neq l_j]$, each $w^{(m)}$ is a learnable weight, and $k^{(m)}$ indicates a Gaussian kernel.

DenseCRF also approximates true distribution $P(\mathcal{L}|\mathcal{X})$ via mean field. It computes a distribution $Q(\mathcal{L}|\mathcal{X})$ that minimize KL-divergence against $P(\mathcal{L}|\mathcal{X})$, which takes the form
\begin{equation}\label{eqn:6}
	Q(\mathcal{L}|\mathcal{X}) = \prod_{i=1}^{N} Q_i(l_i|\mathcal{X}).
\end{equation}

The assumption of mean field approximation is dedicate, that vertices of $G$ are conditional independent from each other. Therefore, minimizing the energy of each node $Q_i$ results in an approximation of global minimum. 

\section{\MakeUppercase{Related Works}}

\subsection{Conventional CRF} 

\begin{figure*}[t]
	\centering
	\includegraphics[width=0.9\linewidth]{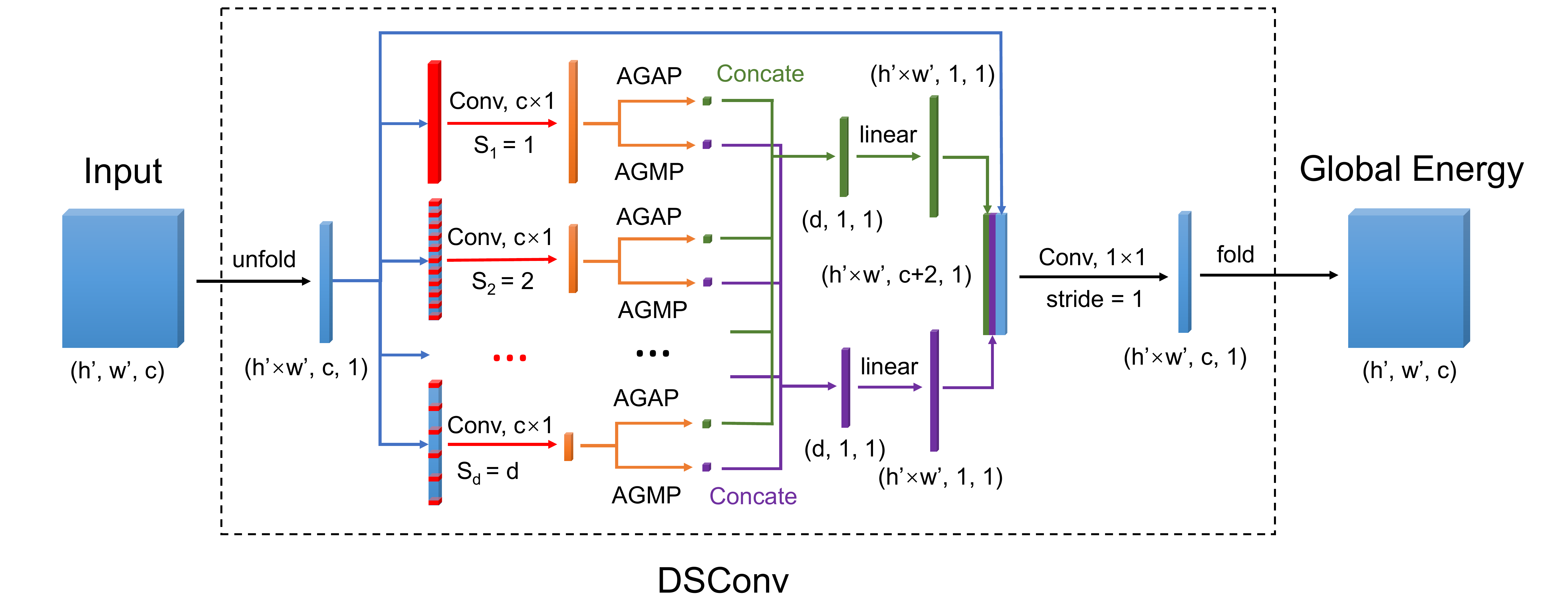}
	\caption{DSConv module. A DSConv module takes down-sampled label volume as the input. Two up-sampled features after linear layers are normalized using Layer Normalization~\cite{ba2016layer}, and activated by ReLU6 function. Concatenation of global pooling artifacts (in green or purple) is the computational bottleneck of proposed DilatedCRF. The folding operation reverses the unfolding of the input feature volume.}
	\label{fig:3}
\end{figure*}

Most of works on CRF are based on mean field approximation theory. For super-pixel image segmentation, Sulimowicz~et~al~\cite{sulimowicz2018superpixel} formulate super-pixel cue onto pairwise potentials, which outperforms DenseCRF; Ma~et~al~\cite{ma2019weakly} introduce two coefficients to unary and pairwise potentials for better encoding super-pixel energy terms; Li~et~al~\cite{li2020weakly} define a third potential to model the cost of assigning super-pixel labels; Yang~et~al~\cite{yang2018high} define a interlayer high order potential to enhance super-pixel segmentation. These works improve DenseCRF w/o adapting CRF to neural networks, which are categorized as conventional CRF. 

\subsection{Refinement for Neural Networks}

As a widely-applied refinement tool for CNN-based approaches, CRF has been proved to be effective in medical image segmentation, according to the references~\cite{zhang2020multiscale, luo2018fast, deng2020deep}.  Lai~et~al~\cite{lai2019deepsqueezenet} integrate DenseCRF into their DeepSqueezeNet for better performance. Zheng~et~al~\cite{zhen2020learning} present a fusion strategy of DenseCRF and attention module for unsupervised video object segmentation. Shimoda~et~al~\cite{shimoda2019self} compare their proposed approach with the refinement output of CRF, and come to the conclusion that their approach is superior, which indicates the failure of CRF in the processing of certain datasets.

\subsection{CRF as Neural Networks}
Recently, several works emphasize on the possibilities of adapting CRF to neural network. CRF-RNN~\cite{zheng2015conditional} is the first architecture that trains a DenseCRF end-to-end via Convolutional Neural Network (CNN) and Recurrent Neural Network (RNN). This work breaks down DenseCRF to common CNN operations except the lattice permutation during message passing of pairwise energy term, and trains these operations iteratively as a RNN. Vemulapalli~et~al~\cite{vemulapalli2016gaussian} propose the Gaussian mean field approximation that computes pairwise energy into three steps - a CNN to generate features, a similarity layer that computes similarity scores, and a matrix generation layer that compute Gaussian weights. Although CNN is well-integrated into the GaussianCRF pairwise energy term, the Gaussian filtering process requires non-standard neural network operations.

Another variant is to compute CRF as a CNN. Teichmann~and~Cipolla~\cite{teichmann2019convolutional} design a ConvCRF that constraints the fully-connected computational graph $G$ into locally-connected blocks via Manhattan distance, which greatly boosts the efficiency of message passing process with the cost of losing long-range connections. Nguyen~et~al~\cite{nguyen2020context} take the advantages of Gated Recurrent Unit~\cite{cho2014learning} (GRU), reformulating CRF-RNN as CRF-GRU that avoids the gradient vanishing and exploding problem of RNN. The gated architecture also enables CRF-GRU to model long-term relationships. Lin~et~al~\cite{lin2016efficient} propose to learn unary and pairwise energies via fully-connected layers from node and edge feature vectors that are yielded from a manually-constructed CRF graph and a convolutional network.  

\begin{figure*}[t]
	\centering
	\centering
	\subfloat[Loss.]{
		\label{fig:4a}
		\includegraphics[width=0.38\textwidth]{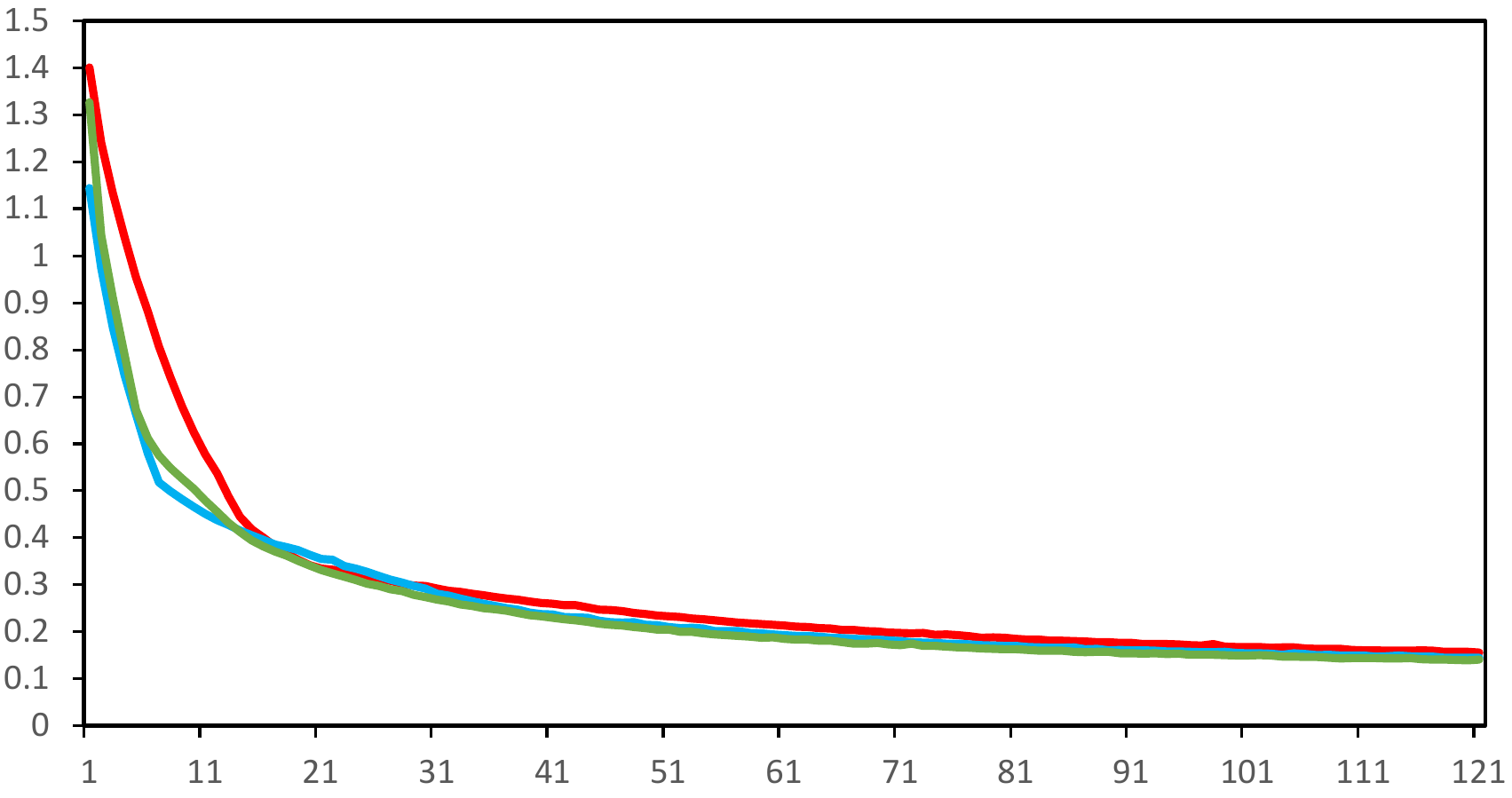}}
	\hspace{15mm}
	\subfloat[Jaccard Index.]{
		\label{fig:4b}
		\includegraphics[width=0.38\textwidth]{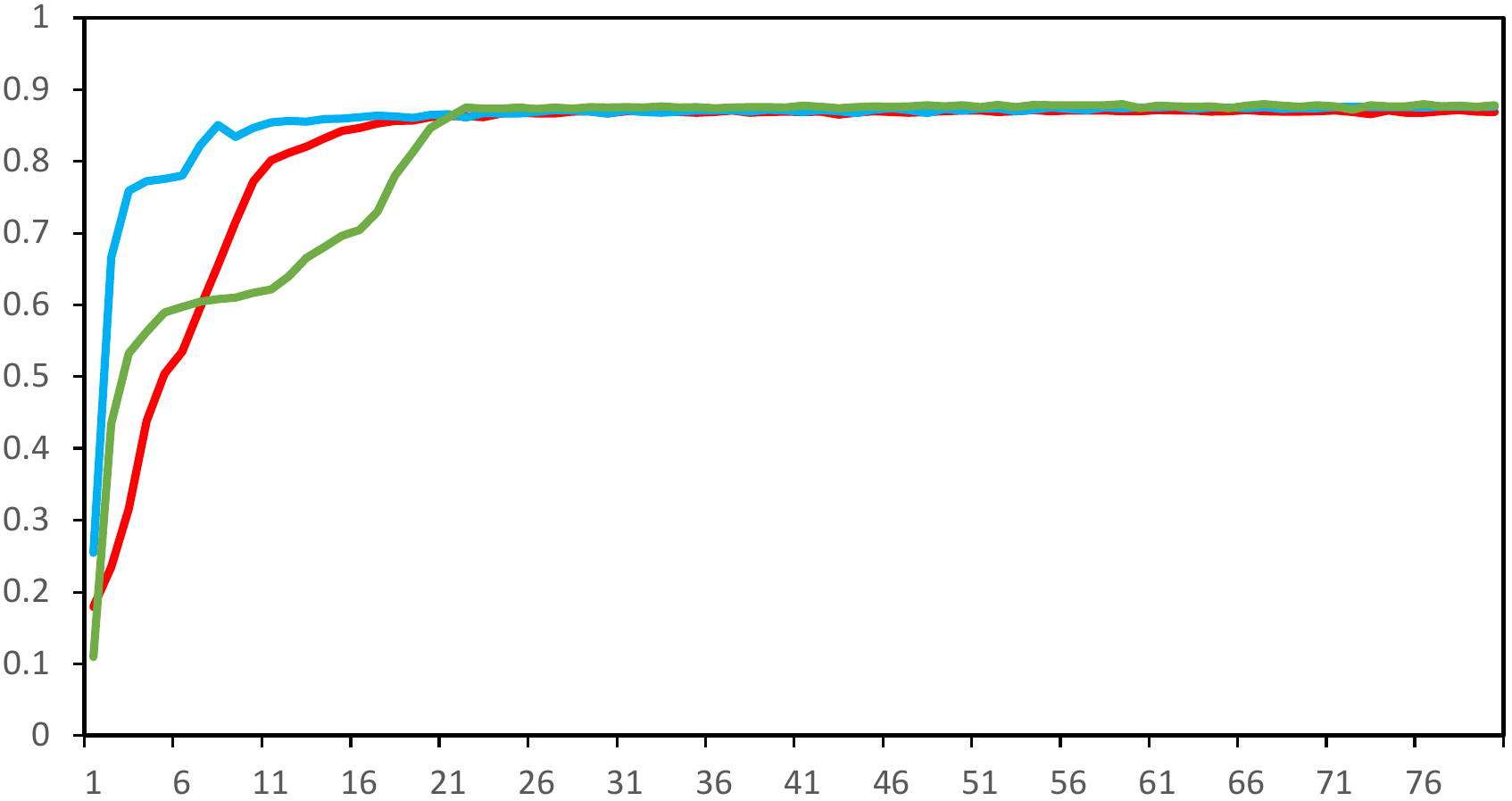}}
	\\
	\subfloat[Precision.]{
		\label{fig:4c}
		\includegraphics[width=0.38\textwidth]{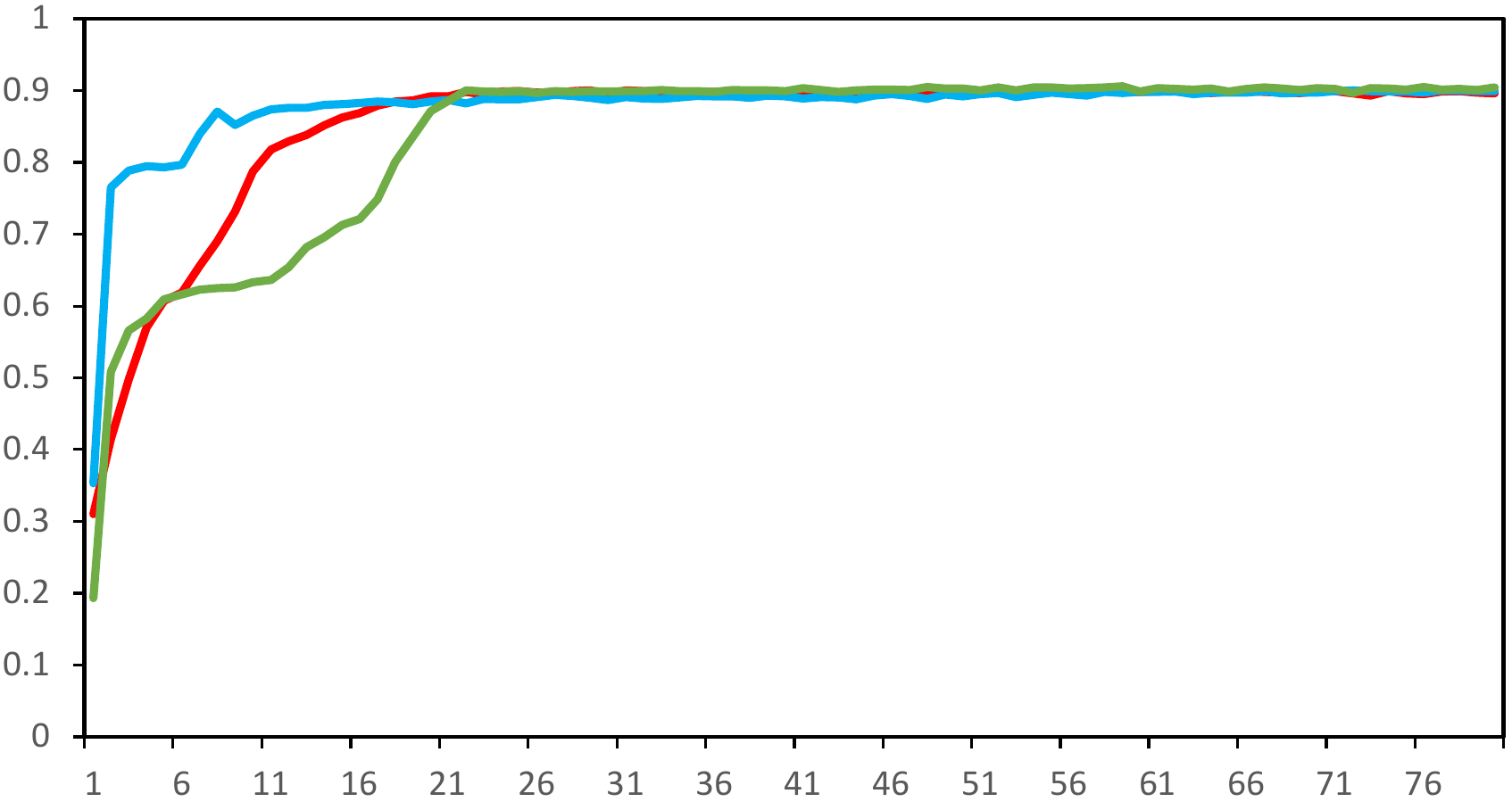}}
	\hspace{15mm}
	\subfloat[Recall.]{
		\label{fig:4d}
		\includegraphics[width=0.38\textwidth]{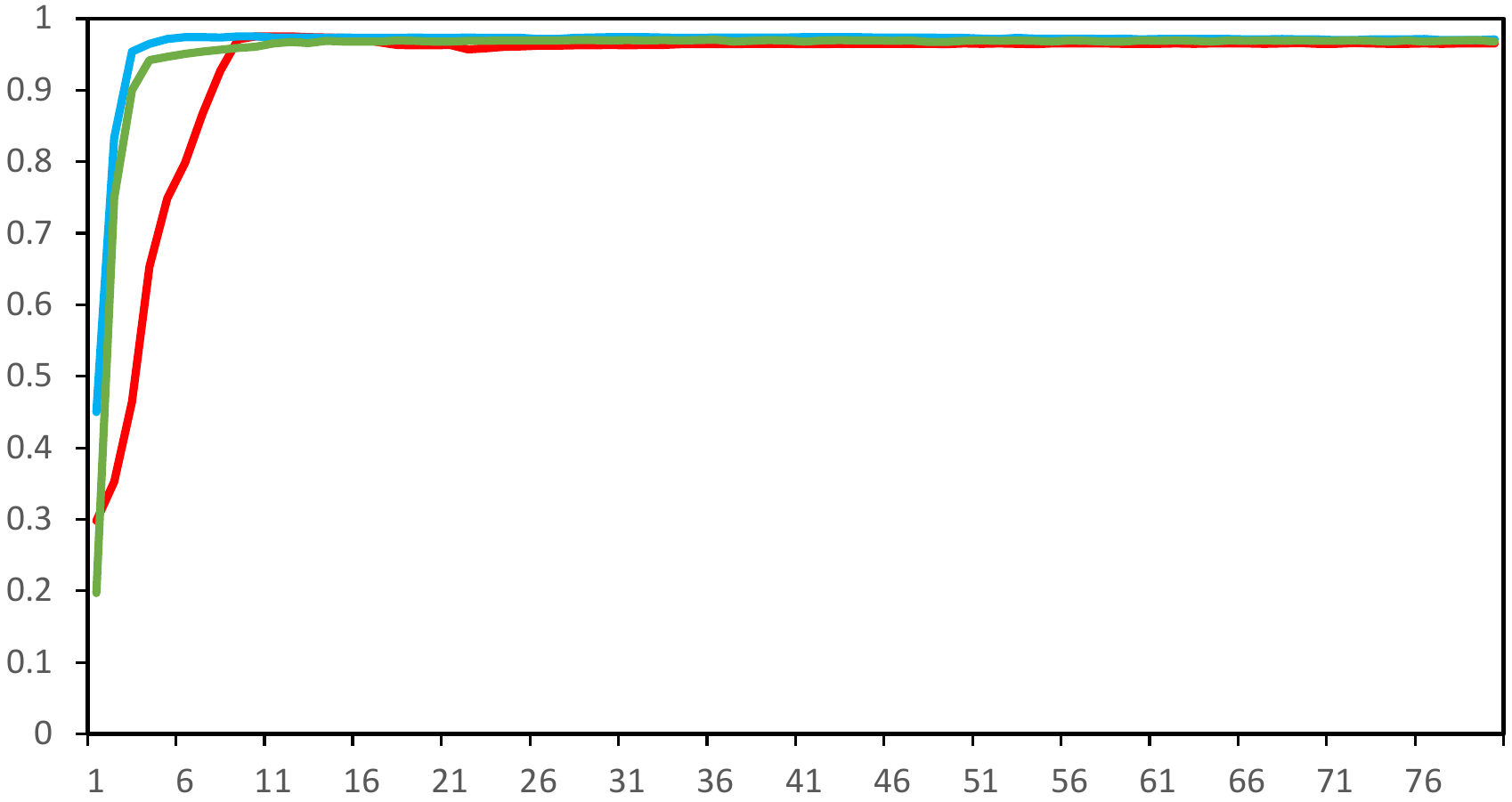}}
	\caption{Training Performance w.r.t. epochs. Red curves: octant-size (60$\times$80), blue curves: quarter-size (120$\times$160), green curves: half-size (240$\times$320). Jaccard Index, precision and recall are calculated by averaging all batches (120 images per batch) of an epoch. Three metrics reach their maximums at epoch 25.}
	\label{fig:4}
\end{figure*}

\footnotetext[4]{Note that $E_{g}(\mathcal{L}|\mathcal{X})$ is not finalized by an activation function after the weighted sum operation.}

\section{\MakeUppercase{Methodology}}

\subsection{Dilated CRF}

Proposed DilatedCRF reformulates Gibbs energy term as fully-learnable unary $E_u(\mathbf{x}_i)$ and global potential\footnotemark[4] $E_g(\mathcal{L}| \mathcal{X})$:
\begin{equation}\label{eqn:7}
	E(\mathcal{L}|\mathcal{X}) = \underbrace{\sum_{1\leq i\leq N}\psi_{u}(\mathbf{x}_i)}_{E_u(\mathbf{x}_i)}+\underbrace{\sum_{1\leq i\leq N}\sum_{j=1}^{2} w_{j}^{(i)}\psi_{j}^{(i)}(\mathcal{L}|\mathcal{X})}_{E_g(\mathcal{L}| \mathcal{X})}.
\end{equation}

In (\ref{eqn:7}), $\psi_{u}(\mathbf{x}_i) = \mathbf{w}_{u}^{T}\mathbf{x}_i$, $\mathbf{w}_u$ is a weights vector with the same length as $\mathbf{x}_i$, which learns the unary term in (\ref{eqn:4}). Particularly, $\mathcal{X}$ is the raw input affordance map, which can either be the artifacts after final softmax layer, or the output using conventional classifiers such as random forest~\cite{yang2011hierarchical}. Unary potential $\psi_u$, global potentials $\psi_1$ and $\psi_2$ can be optimized iteratively during training. $\psi_j$ is defined as
\begin{equation}\label{eqn:8}
	\psi_j(\mathcal{L}|\mathcal{X}) := \mathbf{W}_j\mathbf{f}_{j}(\mathcal{X}),
\end{equation}
where $\mathbf{W}_i$ is a trainable weights matrix that up-samples global feature $\mathbf{f}_i$ to desired length $h\times w$. Moreover, $\mathbf{W}_i$ learns the mapping from $\mathbf{x}$ to potential $\psi_j(\mathcal{L}|\mathbf{x})$ which is not necessarily a valid probability distribution. Follow the DSConv architecture shown in Fig.~\ref{fig:3}, let $d$ be the total number of convolution kernels, and $k$-th kernel has a convolutional stride of $s_k, 1\leq k\leq d$, define $n:=[\frac{N-1}{s_k}+1]$ as the length of output feature, we have
\begin{equation}\label{eqn:9}
	\mathbf{f}_{1}(\mathcal{X})=\frac{1}{n+1}\bigcup_{k=1}^{d}\sum_{\lambda=0}^{n} \mathbf{w}_{(1, \lambda)}^{T}\mathbf{x}_{1+\lambda s_k},
\end{equation}
\begin{equation}\label{eqn:10}
	\mathbf{f}_{2}(\mathcal{X})=\bigcup_{k=1}^{d}\max_{0\leq\lambda\leq n} \mathbf{w}_{(2, \lambda)}^{T}\mathbf{x}_{1+\lambda s_k},
\end{equation}
the operator $\bigcup$ signifies the height-wise concatenation of elements. DilatedCRF is trained end-to-end to minimize KL-Divergence between $\mathcal{L}$ and the ground-truth. All weights are learnable through gradients back-propagation since (\ref{eqn:7}) - (\ref{eqn:10}) are differential.

\subsection{Architecture}

Fig.~\ref{fig:2} illustrates standard operations to perform DilatedCRF inference. We implement a $3\times 3$ fractional max-pooling~\cite{graham2014fractional} layer to down-sample the input label, and nearest interpolation to up-sample refined feature maps. Each channel of the 3-channels unary energy term is computed via $1\times 1$ convolution and batch normalization (BN) layer.

The main contribution of this paper is the DSConv module that encodes the short-term and long-term relationship between feature vectors. As shown in Fig.~\ref{fig:3}, input feature volume is unfolded into a long vector with the shape of $(h'\times w', c, 1)$, sparse convolution using static kernel filtering has $d:=[\sqrt{h'\times w'}/\alpha], \alpha = 2, 10$ strides to perform global correlations, which, in our experiment, we let $\alpha=2$ for octant-size, quarter-size, half-size, and 10 for full-size DilatedCRFs. The AGAP operation (refer to (\ref{eqn:9})) computes average energy intensity, and AGMP operation (refer to (\ref{eqn:10})) locates the strongest response for feature $\mathbf{x}_i$. Both features effectively encode the contribution of $\mathbf{x}_i$ with less parameters than that of implementing two independent linear layers. Additionally, raw input feature is concatenated with the average global feature and maximal global feature for the purpose of reusing features and alleviating the vanishment of gradients. The linear up-sampling operation is based on (\ref{eqn:8}), which is the most critical procedure to learn the relation between $\psi_{j}^{(i)}(\mathcal{L}|\mathcal{X})$ and $\mathbf{x}_i$. Equation (7) is realized by performing $1\times 1$ convolution upon concatenated features.

\begin{table*}
	\renewcommand{\arraystretch}{1.4} 
	\begin{center}
		\setlength{\tabcolsep}{5.5mm}{ 
			\begin{tabular}{lccccc}
				\whline
				Items & Full-Size & Half-Size & Quarter-Size & Octant-Size & LR-ASPP~\cite{howard2019searching}\\\hline
				Total parameters (M) & {\bf 0.09} & 21.66 & 2.77 & 0.36 & - \\
				Inference time (\textit{ms}) & 13.99 & 159.43 & 18.81 & \bf 8.91 & - \\\shline
				Accuracy (\%) & 96.56 $\pm$ {\bf 2.76} & 96.22 $\pm$ {\bf 2.76} & 96.35 $\pm$ 2.78 & 96.29 $\pm$ 2.71 & {\bf 96.61} $\pm$ 2.80\\
				Jaccard Index (\%) & 40.75 $\pm$ 23.42 & 40.77 $\pm$ 23.20 & {\bf 41.21} $\pm$ 23.30 & 39.84 $\pm$ {\bf 22.67} & 40.07 $\pm$ 23.72\\
				Precision (\%) & 57.37 $\pm$ 28.12 & 53.09 $\pm$ 27.68 & 54.30 $\pm$ {\bf 27.58} & 54.67 $\pm$ 28.23 & {\bf 58.16} $\pm$ 28.20 \\
				Recall (\%) & 58.47 $\pm$ 24.78 & {\bf 64.42} $\pm$ 23.74 & 62.84 $\pm$ 23.85 & 61.11 $\pm$ {\bf 23.17} &  56.02 $\pm$ 25.76 \\
				Dice Coefficient (\%) & 53.94 $\pm$ 24.24 & 54.03 $\pm$ 24.01 & {\bf 54.45} $\pm$ 24.08 & 53.20 $\pm$ {\bf 23.69} & 53.07 $\pm$ 24.86\\
				\whline
		\end{tabular}}
	\end{center}
	\caption{DilatedCRF evaluation on test set, all metrics are formatted as $\mathrm{mean}\pm\mathrm{std}$. Our training samples are derived from state-of-the-art LR-ASPP~\cite{howard2019searching}, and average inferential runtime over test set is reported. Inference time of LR-ASPP is not presented because we only focus on the CRF-based post-processing.}
	\label{tab:1}
\end{table*}

\subsection{Training}

Assume the true conditional probability distribution of ground-truth is denoted as $G(\mathcal{L}^*|\mathcal{X})$, KL-Divergence is calculated by
\begin{equation}
	\mathrm{KLD}(G||P) = \sum_{\mathcal{L}}{G(\mathcal{L}^*|\mathcal{X})}\log\frac{G(\mathcal{L}^*|\mathcal{X})}{P(\mathcal{L}|\mathcal{X})}.
\end{equation}

If $G$ is one-hot encoded, KL-Divergence equals the value of cross-entropy loss (CELoss), we thereby utilize a weighted form of CELoss function for training DilatedCRF:
\begin{equation}
	\mathrm{CELoss}(l, l^*) = -\sum_{cls} (\beta_{cls}\sum_{cls} \log l),
\end{equation}
where $\beta_{cls}$ is a manually-set weight for the category $cls$, and the subscript variable $cls\in\{0, 1, 2\}$ indicates a category of \{`background', `negative samples', `valid suckable area'\}.

We trained the network by setting $\beta$ to (0.25, 0.25, 0.5), the weight assigned for category 2 is larger than other weights because the valid suckable regions usually occupy a small portion of the image, while we only need to evaluate this category. The suction dataset~\cite{suctiondataset} has a training split of 1470 images, and a testing split of 367 images, each image has a resolution of 480$\times$640.

\section{\MakeUppercase{Experiments}}\label{sec:exp}

\subsection{Implementation Details}

\subsubsection{Training and test}

We follow a four-stages strategy to train and test DilatedCRF w/ a RTX3080 GPU:

\begin{itemize}
	\item{Pre-train classifiers on the training set.}
	\item{Generate predicated post-softmax\footnotemark[5] affordance maps for both training and test sets.}
	\footnotetext[5]{Alternatively, pre-softmax affordance maps are also theoretically valid.}
	\item{Train DilatedCRF using the predicted affordance maps w/ AdamW optimizer (w/o weight decay) and a learning rate of 1$\times 10^{-3}$.}
	
	\item{Test the DilatedCRF using a selected checkpoint, and evaluate results by segmentation metrics.}
\end{itemize}

Typically, it takes around 5 hours to train a half-size DilatedCRF. We quantitatively evaluate the results by Jaccard Index, Precision, Recall, Dice coefficient, and Accuracy, which are defined as
\begin{equation}
	\mathrm{\mathbf{Jaccard} = \frac{TP}{FP + FN + TP}},
\end{equation}
\begin{equation}
	\mathrm{\mathbf{Accuracy} = \frac{TP + TN}{FP + TP + FN + TN}},
\end{equation}
\begin{equation}
	\mathrm{\mathbf{Precision} = \frac{TP}{FP + TP}, ~\mathbf{Recall} = \frac{TP}{TP + FN}},
\end{equation}
\begin{equation}
	\mathrm{\mathbf{Dice} = \frac{2\cdot Precision\cdot Recall}{Precision + Recall}}.
\end{equation}

\subsubsection{Performance in Training}

In Fig.~\ref{fig:4}, we report more training details for half-size, quarter-size and octant-size DilatedCRFs. Over-fitting is not observed as loss curves (in Fig.~\ref{fig:4a}) converge at around 0.11, and curves of metrics arrive at their maximums before the convergence.

It can also be inferred from Fig.~\ref{fig:4b}-\ref{fig:4d} that quarter-size DilatedCRF manifests the fastest convergence speed during training, and Jaccard Index curves behaves extremely similar to precision curves, which indicates precision is more decisive than recall for the suction dataset. Although differ in sizes, all three models converge to the same numerical level upon the training set. 

\begin{table}
	\renewcommand{\arraystretch}{1.4} 
	\begin{center}
		\setlength{\tabcolsep}{4,5mm}{ 
			\begin{tabular}{lcc}
				\whline
				Items & ConvCRF & DenseCRF\\\hline
				Inference time (\textit{ms}) & 14.10 & 450.16\\\shline
				Accuracy (\%) & 96.40 $\pm$ {\bf 2.84} & {\bf 96.74} $\pm$ 2.89\\
				Jaccard Index (\%) & {\bf 34.35} $\pm$ {\bf 22.95} & 33.76 $\pm$ 25.63\\
				Precision (\%) & 54.23 $\pm$ {\bf 29.17} & {\bf 60.49} $\pm$ 32.35\\
				Recall (\%) & {\bf 48.76} $\pm$ {\bf 28.43} & 43.31 $\pm$ 30.74\\
				Dice Coefficient (\%) & {\bf 46.77} $\pm$ {\bf 25.94} & 45.05 $\pm$ 28.75\\
				\whline
		\end{tabular}}
	\end{center}
	\caption{Performance of ConvCRF and DenseCRF on test set. The metrics are formatted as $\mathrm{mean}\pm\mathrm{std}$. We implement the same LR-ASPP unary as illustrated in Tab.~\ref{tab:1}.}
	\label{tab:2}
\end{table}

\subsubsection{Test performance}\label{sec:test}

As shown in Tab.~\ref{tab:1}, all DilateCRFs outperform the original LR-ASPP~\cite{howard2019searching} approach on Recall and Dice scores, which indicates the potential of DilatedCRF in locating more objects, while maintaining high precision and lower uncertainties manifested by standard deviation. Octant-size DilatedCRF has the coarsest affordance predictions for the 8$\times$ lower down-sampling, however, it shows competitive performance both in Accuracy and Jaccard Index, and only takes additional 9\textit{ms} inference runtime for refinement. The quarter-size DilatedCRF performs the best among all DilatedCRFs considering the trade-off between inference time and all metrics. Furthermore, it achieves the highest Dice score. Half-size DilatedCRFs continue to robustly improve Recall score whereas the twofold enlargements of image height and width greatly slows down inference procedure, bringing in more noisy features. For full-size DilatedCRF w/o the down-sampling and up-sampling layers shown in Fig.~\ref{fig:2}, coefficient $\alpha$ is set to 10, which dramatically boosts the runtime efficiency.

\subsection{Compare to other CRFs}

In this section, we report our quantitative and qualitative results with ConvCRF~\cite{teichmann2019convolutional} and DenseCRF~\cite{krahenbuhl2011efficient}. ConvCRF is tested w/ a RTX3080 GPU, and DenseCRF is tested w/ a 3.60GHz Intel i9-10850K CPU. 

\begin{figure*}[t]
	\centering
	\includegraphics[width=\linewidth]{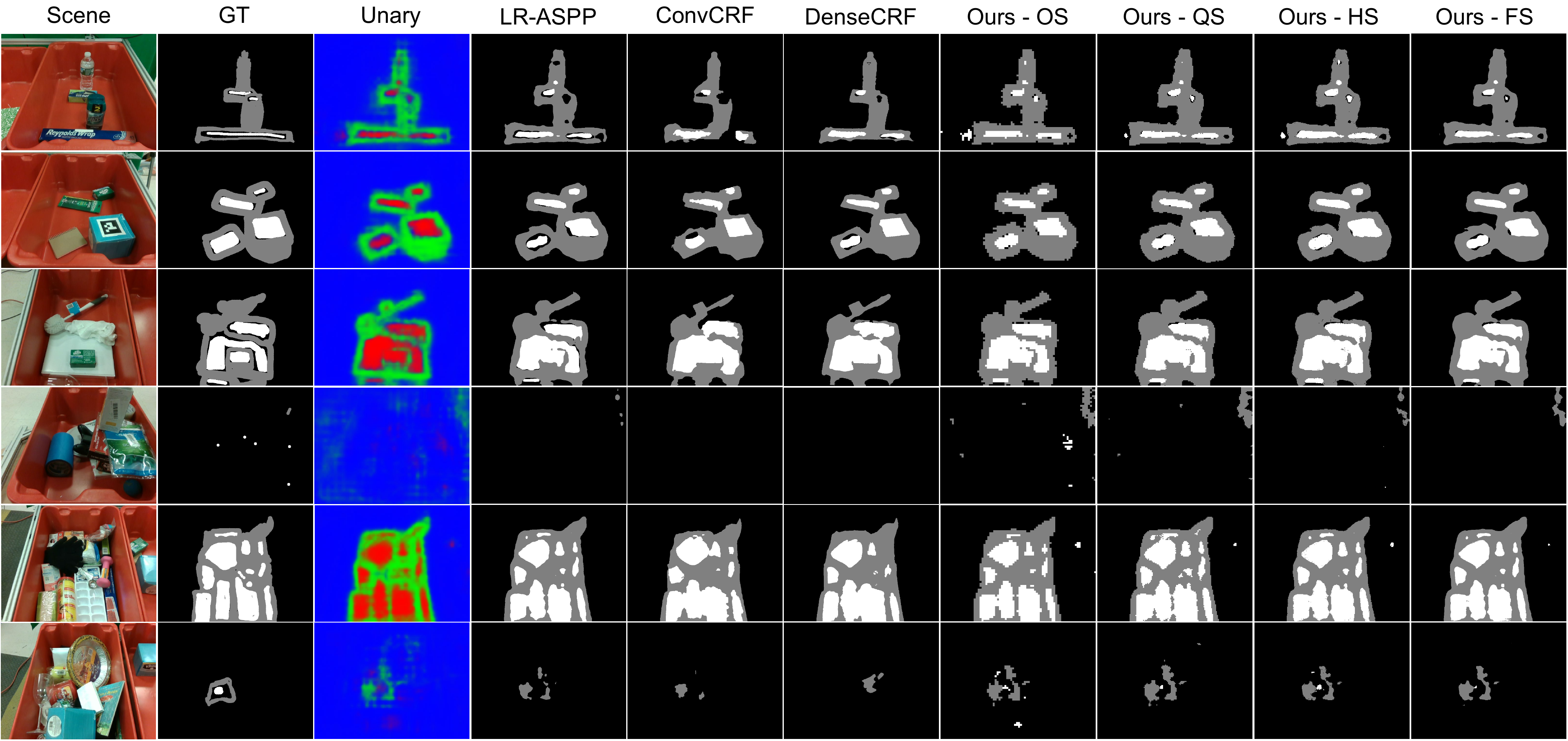}
	\caption{Visualization of affordance maps. Black, grey, white regions correspond to background, negative samples, suction areas respectively. `OS' - octant-size DilatedCRF, `QS' - quarter-size DilatedCRF, `HS' - half-size DilatedCRF, and `FS' - full-size DilatedCRF.}
	\label{fig:5}
\end{figure*}

\subsubsection{Quantitative Evaluation}

As shown in Tab.~\ref{tab:2}, DenseCRF achieves the highest average precision $60.49\%$ and accuracy $96.74\%$ among all tested approaches. However, due to the largest variances, DenseCRF does not perform as stably as desired. Also, its lower recall and Jaccard index significantly degrade its performance on test set. ConvCRF, in the other hand, make a relatively better trade-off between recall and precision. It is noteworthy that, although ConvCRF has an average performance, it runs 32$\times$ faster than DenseCRF.

\subsubsection{Qualitative Results}

Fig.~\ref{fig:5} depicts affordance maps, images of scenes, and post-softmax unary maps. Both ConvCRF and DenseCRF enhance boundaries better than neural network based approaches. Nonetheless, the defects of these two approaches are obvious. First, they tend to filter out small regions, which may lower the success rate of suction-type picking. Second, the relationships between separated regions may not being well-encoded due to the chromatic continuities of raw scenes, which results in largely-connected affordance maps.

In comparison with ConvCRF and DenseCRF, the proposed approaches better recognize the relationships between separated regions, and locate more objects whereas some of them are not from the ground-truth. Although these predictions are categorized as false negatives, as depicted in Fig.~\ref{fig:5}, a robotic hand may successfully pick an object from bins under the guidance of these predictions. Generally, the proposed approaches directly learn vanilla features from unary maps, progressively refine memorized features towards the ground-truth. Furthermore, we test the checkpoint after 200 epoches, and discover a similar performing trend as of DenseCRF, i.e., higher accuracy and precision but lower recall and Jaccard index than LR-ASPP artifacts, which indicates the potential of DSConv module to approximate the global optimum of fully-connect CRF. 

\section{\MakeUppercase{Conclusion and Future Works}}

We propose DilatedCRF to approximate fully-connected CRF without relying on mean-field approximation. DilatedCRF reformulate unary and pairwise energy terms as learnable unary and global energy terms. The reformulation adapts the end-to-end neural networks to conventional CRF without CPU-only operations, which is faster and more liable for real-world implementations. To realize this reformulation, we design a dilated sparse convolution module that takes advantages of adaptive global average-pooling and adaptive global max-pooling, as well as the 3-stages DilatedCRF pipeline to accommodate varieties of unary maps. Extensive experimental results show the high efficiency, and competitive performance of the DilateCRF pipeline for realtime robotic suction-type bin-picking. 

Our future works include: 
\begin{itemize}
	\item{Apply highly-precise backbone networks to better learn the unary terms.}
	\item{Use network adaptive searching strategies to find the best strides for DSConv module.}
	\item{Integrate high-order contextual information into global energy terms, this may further enhance the regional encoding capabilities of DilatedCRF.}
	\item{Speed up the inferential procedures of full-size and half-size models by reducing concatenation operations.}
\end{itemize}

\addtolength{\textheight}{-12cm}  

\section*{ACKNOWLEDGMENT}

We thank Li, Rui for the sponsorship of NVIDIA GPUs and INTEL CPUs, and M. Teichmann, R. Cipolla for providing open source codes of their~\cite{teichmann2019convolutional} invaluable works.

\bibliographystyle{IEEEtran} 
\bibliography{egbib}

\end{document}